\documentclass{llncs} % 
\usepackage{times}
\usepackage{helvet}
\usepackage{courier}
\usepackage[english]{babel}
\usepackage{hyperref}
\usepackage{amssymb}
\usepackage{amsmath}
\usepackage{verbatim}

\newcommand{\wdt}[1]{\textsf{\small #1}}

\newcommand{\tr}[1]{{\color{orange} #1}} 
\newcommand{\trs}[2]{#1 {\color{pink} #2}} 

\renewcommand{\tr}[1]{#1} 
\renewcommand{\trs}[2]{#2}

\tr{\pagestyle{plain}}
\begin{document}

\title{Wikidata on MARS\trs{}{\thanks{This is an extended version of a paper that appears in the 2020 Description Logic Workshop.}}}

\author{Peter F. Patel-Schneider \inst{1} \and David Martin \inst{2}} 
\institute{Palo Alto Research Center \and Unaffiliated} 

\maketitle
\begin{abstract}
Multi-attributed relational structures (MARSs) have been proposed as a formal data model for generalized property graphs, along with multi-attributed rule-based predicate logic (MARPL) as a useful rule-based logic in which to write inference rules over property graphs. Wikidata can be modelled in an extended MARS that adds the (imprecise) datatypes of Wikidata. The rules of inference for the Wikidata ontology can be modelled as a MARPL ontology, with extensions to handle the Wikidata datatypes and functions over these datatypes. Because many Wikidata qualifiers should participate in most inference rules in Wikidata a method of implicitly handling qualifier values on a per-qualifier basis is needed to make this modelling useful. The meaning of Wikidata is then the extended MARS that is the closure of running these rules on the Wikidata data model. Wikidata constraints can be modelled as multi-attributed predicate logic (MAPL) formulae, again extended with datatypes, that are evaluated over this extended MARS. The result models Wikidata in a way that fixes several of its major problems.

\end{abstract}

\section{Introduction}

Wikidata is already an extremely useful source of information about notable
entities in the world, containing high-quality information about over 87
million items.

However, Wikidata must be used with care as it does not present a cohesive
view of the world.  To pick a simple example, Wikidata has a property for
\wdt{spouse}\footnote{For readability we use the English label (with underscores) to identify Wikidata items, here {\small \url{https://www.wikidata.org/wiki/Property:P26}}, \tr{even though that might not uniquely identify an item.}}
Wikidata also states
that \wdt{spouse} is its own inverse and has symmetric constraints among its
property constraints.  Either of these alone should be adequate to allow
users of Wikidata to assume that to find the spouse of someone it suffices
to query that someone's value for spouse.  But this is not necessarily
so---as of mid-June 2020 there were over thirty-eight hundred non-symmetric spousal
relationships in Wikidata, for items ranging from Roman gods to
politicians.\tr{\footnote{The number of violations is helpfully provided on a database report page that collects symmetry violations for spouse.}}
So any attempt to accurately determine the spouse of someone in Wikidata
needs to perform two queries.  Making this fix for spouse is relatively
easy, but in general it is impossible for an application to automatically
determine what to do to access all the information that it needs that is
inherent in Wikidata.    

Wikidata has a large, but simple, ontology, containing classes such as \wdt{human} and \wdt{person} and a \wdt{subclass\_of} relationship from \wdt{human} to \wdt{person}.  Retrieving all the people in Wikidata should thus be a simple matter of finding all the items that have an \wdt{instance\_of} link to \wdt{person}.  But as of mid-June 2020 there are only about 30 items that have an \wdt{instance of} link to \wdt{person}---far fewer than the over 7.9 million humans in Wikidata.  To find all instances of \wdt{person} in Wikidata requires asking for instances of the transitive closure of \wdt{subclass\_of} under \wdt{person}.
Again, it is not hard to remedy this particular problem, and it is so important that most accessors of Wikidata make the extra effort to look through subclasses.  On the other hand, based on our personal experiences the analogous workaround for subproperties is often not done.

As yet another illustrative example, what if an
application wants to know what information is current, i.e., true at the
present time.   Wikidata has quite a number of constructs that specify
temporal extents for pieces of information, including \wdt{end time}.  How can the application know
what it needs to do to filter out statements that are not valid at the current
time?

Providing an interface to Wikidata that presented all and only the valid
information would make it much easier to write applications that just want
to know what's true based on what is in Wikidata, as opposed to how the
information is said or who said it.

Maybe this effort is doomed to be fruitless.  Maybe Wikidata is just
data and there is no way to provide a decent logical view of the data in
Wikidata.  This would indeed be a shame, as Wikidata is the best example of
an open Knowledge Graph, and so it should be possible to treat the Wikidata
data logically.

The need for well-founded reasoning capabilities in Wikidata has been noted and lightly documented in a 
WikiProject\tr{\footnote{https://www.wikidata.org/wiki/Wikidata:WikiProject\_Reasoning}}, but to date this project has not produced any concrete results.  The current work may be viewed as a proposal to satisfy a superset of the requirements documented there.
\vspace*{-1ex}
\subsection{Areas To Be Addressed}
\vspace*{-1ex}
Wiki{\em data} is data, but many portions of Wikidata that are vital to
Wikidata's utility and success 
are conceived, discussed, and documented as if Wikidata is more than just data,
as illustrated by \wdt{spouse} and \wdt{instance\_of} above.
To reliably provide construction of the meanings implicit in the
descriptions of even the central properties of Wikidata requires a formal theory
for the properties and an implementation of this theory.
Without such, different users and tool builders end up determining and
implementing the meaning on their own, and this often produces
different, non-compatible results.

One way of providing a common meaning is via a set of axioms in a logic and
a service that implements reasoning in the logic, either in general or for
the specific axioms.  The service then provides a common resource that
serves up the meaning of the various pieces of Wikidata.
This is particularly vital for the central, ontological portion of
Wikidata, including \wdt{instance\_of} as well as \wdt{symmetric\_property}.

Other parts of Wikidata are also currently under-specified and
differentially implemented and can be improved by a logical formulation.
Wikidata
constraints\tr{\footnote{https://www.wikidata.org/wiki/Help:Property\_constraints\_portal}}
express regularities that are supposed to hold in general, but which may
have exceptions.  These constraints are not used to draw inferences but instead to
point out potential problems to interested constributors who can then either
fix the problem or indicate that the particular anomaly is acceptable.

Wikidata constraints, albeit useful, are represented and processed in an incomplete, ad hoc fashion.  Although in most cases they are declared and documented reasonably clearly, the declarations do not fully express their meaning, and thus do not provide a precise, unambiguous basis, or a logical foundation, for constraint-checking implementations.  Further, building a constraint checker for a new constraint may be laborious, idiosyncratic, and error-prone.  A logical formulation and implementation of constraints would permit constraints to be quickly specified and eliminate the implementation burden for each new type of constraint.

As noted in \cite{marx2017logic}, Wikidata’s custom data model supports attributed statements (with the attributes referred to as \emph{qualifiers}), and allows attributes with multiple values.  \cite{marx2017logic} refers to this sort of generalized Property Graph model as a \emph{multi-attributed graph}, and observes that ``In spite of the huge practical significance of these data models ..., there is practically
no support for using such data in knowledge representation.''  This creates not only the challenge of constructing a logic in which qualifiers are first-class citizens, but also the challenge of creating a reasoning framework in which the behavior of qualifiers in axioms can be specified in a practical fashion. 

Wikidata has many classes and properties, such as \wdt{female\_human}, that are missing many of their expected instances or relationships.
Axioms that give a logical formulation for these classes and properties would add back their missing elements and eliminate the incorrect information that currently comes from  accessing them in Wikidata.

Wikidata has several ways of providing something other than a particular value for a property.
The \emph{some-value snak} represents some ``unknown'' value and the \emph{no-value snak} says that there is no value.
However the precise meaning of both of these is unclear, 
and a logical formulation for them  would provide a better meaning for these parts of Wikidata.
\vspace*{-1.5ex}
\subsection{Contributions}
\vspace*{-.5ex}
We provide a logical foundation for Wikidata, building on previous work on 
multi-attributed predicate logic (MAPL), including 
multi-attributed rule-based predicate logic (MARPL) and
multi-attributed relational structures (MARSs)
\cite{marx2017logic}.
Our work
%\begin{itemize}
%    \item
1) supports the explicit expression of Wikidata's ontological axioms, and their use in accessing Wikidata;
%    \item 
2) supports the explicit expression of other axioms that can benefit other areas of Wikidata;
%    \item
3) supports the specification of qualifiers as first-class citizens in axioms and constraints,
and provides a means for effectively handling them;
and
%    \item
4) supports the expression of nearly all current Wikidata property constraints, plus a variety of other constraints and the identification of constraint violations,
%\end{itemize}
all in one logical framework.

\section{Development}

\label{subsec:marsstuff}
Wikidata's custom data model goes
beyond the {\it Property Graph} data model, which associates sets
of attribute-value pairs with the nodes and edges of
a directed graph\tr{, by allowing for attributes
with multiple values}.  Marx {\em et al.}  refer to such generalized
Property Graphs as {\it multi-attributed graphs}, and observe that
``In spite of the huge practical significance of these data models
..., there is practically no support for using such data in knowledge
representation''.  Given that motivation, Marx {\em et al.} introduce the
{\it multi-attributed relational structure} (MARS) to provide a formal
data model for generalized Property Graphs, and {\it multi-attributed
  predicate logic} (MAPL) for modeling knowledge over such structures.
MARS and MAPL may be viewed as extensions of FOL to support the use of
attributes (with multiple values).

The essential new elements over FOL are these:
\begin{itemize}
\item
  a {\it set term} is either a set variable or a set of
  attribute-value pairs $\{a_1 : b_1, ..., a_n : b_n\}$, where $a_i,
    b_i$ can be {\it object terms}.  Object terms are the usual
  basic terms of FOL, and can be either constants or {\it object
    variables}.
  \item
a {\it relational atom} is an expression $p(t_1, ... t_n)@S$, where
$p$ is an {\it n-ary} predicate, $t_1, ... t_n$ are object terms and $S$
is a set term.
\item
  a {\it set atom} is an expression $(a : b) \in S$, where $a, b$ are
  object terms and $S$ a set term.
\end{itemize}

These elements are best illustrated with a simple example (taken directly from 
\vspace*{-1ex}
\cite{marx2017logic}):
\begin{equation}
\label{eqn:spouseRule}
\begin{aligned}
  \forall x,y,z_1,z_2,z_3 . 
\wdt{spouse}(x, y)@\{\wdt{start} : z_1, \wdt{loc} : z_2, \wdt{end} : z_3\} \\
\;\;\;\; \rightarrow \wdt{spouse}(y, x)@\{\wdt{start} : z_1, \wdt{loc} : z_2, \wdt{end} : z_3\}
\end{aligned}
\end{equation}
\vspace*{-1ex}

\noindent
This MAPL formula states that \wdt{spouse} is a symmetric relation,
where the inverse statement has the same start and end dates, and
location.  The entire second line of the formula is a relational atom,
which includes the set term $\{\wdt{start} : z_1, \wdt{loc} : z_2,
\wdt{end} : z_3\}$.  If that set term were represented by a set
variable $U$, then one could make an assertion about its membership
using the set atom $(\wdt{start} : z_1) \in U$.

\tr{In Wikidata terms, this particular relational atom (once $x$ and $y$
have been instantiated to specific Wikidata items) corresponds to a
statement, and each attribute-value pair (once the $z_i$ variable has
been instantiated to a specific value), corresponds to a qualifier of
the statement.  ($x$, of course, is called the subject of the
statement, and $y$ the value or object of the statement.)  While MAPL
allows for predicates of arbitrary arity, in Wikidata all statements
are triples; i.e. Wikidata properties have arity 2.
}

Marx {\em et al.} go on to introduce multi-attributed rule-based
predicate logic (MARPL), roughly the Horn-clause fragment of MARS plus functions that compute the attributes of atoms in the consequent of rules.
MARPL is decidable for fact
entailment, but still provides a high level of expressivity.
Note that Formula
\ref{eqn:spouseRule} falls within the MARPL fragment.

\subsection{eMARS}

MARPL is close to what is needed as a basis for Wikidata but MARPL is missing
datatypes and each rule in MARPL has to specifically account for attributes
(essentially the qualifiers of Wikidata).
Datatypes play a large role in Wikidata and handling Wikidata qualifiers correctly requires accounting for many qualifiers in each of many rules, which is infeasible from a practical perspective with MARPL.
Using an extension of MARPL means that we have the chance of being able to relatively easily compute the consequences of Wikidata.
We also use MAPL formulae in constraints, so we also need to extend MAPL.

As MARPL and MAPL are based on MARSs
we start by adding datatypes to MARSs.
As Wikidata, like RDF \cite{rdf}, has a single domain for everything, including predicates, 
we will be extending MARS in this direction, but in a way that does not permit some of the strange situations possible in RDF.
Because of space limitations we write our definitions in a somewhat informal manner, but the details can be filled out as is usual.

\begin{definition}
A datatype theory, $\mathcal{D}$, consists of a finite set of named datatypes, $D$, each of which has a finite or infinite set of data values; a finite set of named and typed datatype relations, $R$, over $D$; and a finite set of named and typed datatype functions, $F$, over $D$.  The relations are closed under negation.
\end{definition}

\tr{So datatype theory for the rationals and integers would have as data values all the rational numbers (with the integers as a subset).  The datatype functions and relations could include the comparison relations (both within each datatype and between the two datatypes) and arithmetic functions.}

\begin{definition}
An extended MARS (eMARS), $\mathcal{M}$, is a MARS 
extended with a datatype theory, $\mathcal{D}$.
All datatypes, datatype relations, and datatype functions of $\mathcal{D}$ 
as well as all predicates are distinct elements of the domain of $\mathcal{M}$, $\delta^{\mathcal{M}}$.   
All data values in $\mathcal{D}$ are also elements of $\delta^{\mathcal{M}}$.
Each datatype is a unary predicate of $\mathcal{M}$ which is true on the data values of the datatype.
\end{definition}
\tr{The domain elements for datatypes, datatype relations, datatype functions, and predicates are all distinct, thus eliminating several unusual situations that can occur in RDF and can be forced in extensions of RDF.}

\vspace*{1ex}

\noindent
\textbf{Wikidata datatypes.}
The defined datatypes of Wikidata\footnote{Most information shown here about the Wikidata data model is taken from {\small \url{https://www.mediawiki.org/wiki/Wikibase/DataModel}}.} are 
\wdt{IriValue};
\wdt{StringValue};
\wdt{MonolingualTextValue}, strings with language tags;
\wdt{MultilingualTextValue}, strings in multiple langauges;
\wdt{QuantityValue}, with an associated unit;
\wdt{GeoCoordinatesValue}, in some coordinate system;
and
\wdt{TimeValue}, including a timezone.
The last three of these are unusual in that they have imprecise values, containing a main value and some notion of precision (not necessarily symmetric), indicating an interval or range of possible values.

In our datatype theory for Wikidata all datatypes have equality and inequality relations that take into account all aspects of the data values.  
The precise datatypes have the usual set of relations and functions that extract their pieces.
\tr{
\wdt{IriValue} has no extra relations or functions.
\wdt{StringValue} has as well the usual lexicographic relations and pattern matching relations.
The relations for \wdt{StringValue} are extended to \wdt{MonolingualTextValue} (the languages have to be the same for the relation to hold).  There are functions to extract the string and language tag from \wdt{MonolingualTextValue}.
The relations for \wdt{StringValue} are extended to \wdt{MonolingualTextValue} by requiring that the strings for each languages be in the relation.  There is a function to extract the string for a language tag, which returns the empty string if there is no string with that language tag in the data value.
}

Imprecise datatypes have as well equality and inequality for their main values,
a predicate for whether a value is precise,
and overlaps and disjoint and emptiness.
They also have intersection functions that result in a smallest (in imprecision) value that includes the intersection of the two values, if the two values intersect, otherwise an empty value (not currently in Wikidata) results;
and a kind of union function that produces a smallest (again in imprecision) value that includes both the values.
(Some imprecise Wikidata datatypes have multiple values that are equivalent.   Some imprecise Wikidata dataypes are not closed under intersection.)

\wdt{QuantityValue} has ordering relations of three flavours,
one for the main value, e.g., the main value of the first argument is less than that of the second;
one for necessary ordering, e.g., all possible values of first argument are less than those of the second;
and
and one for possible ordering, e.g., some possible values of first argument is less than one of the second.
\wdt{GeoCoordinatesValue} has various direction relations, such as
\wdt{north\_of}, \wdt{must\_be\_north\_of} and \wdt{can\_be\_north\_of},
the latter two taking into account imprecision.

\wdt{TimeValue} has ordering relations, as for \wdt{QuantityValue}, although the base names are \wdt{before} and \wdt{after}.
There are also functions to return the part of one time that must (or can) be before (or after) the second.
As well, there is a \wdt{first} function whose returned \wdt{TimeValue} has a main time, first possible time, and last possible time that are the earlier ones from either of its two (\wdt{TimeValue}) arguments, and an analogous \wdt{last} function.

This is just an initial set of relations and functions, and likely needs expansion.
\tr{There are currently no construction functions, for example, to construct a 
\wdt{MultilingualTextValue} from several \wdt{MonolingualTextValue}s, pending an analysis of which construction functions do not cause computational difficulties.}

\subsection{eMAPL}

Extended MAPL (eMAPL) is then MAPL based on eMARS.
Datatype predicates and functions are allowed in eMAPL, of course.
To support the representation of constraints,
we add equality and, as syntactic sugar, counting quantifiers.

\begin{definition}
  eMAPL terms are MAPL terms with the addition of datatype function applications.
  eMAPL formulae are MAPL formulae using eMAPL terms plus the addition of
datatype relations as predicates, an equality predicate, and counting quantifiers.
  Atoms with datatype relations cannot have non-empty attribute sets.
\end{definition}

\begin{definition}
  An eMAPL interpretation is a MAPL interpretation with the following additions and modifications:
  \begin{itemize}
    \item An eMAPL interpretation is into an eMARS.
    \item The interpretations of datatypes, predicate and datatype relations, and functions are themselves.
    \item The interpretation of a datatype function term is the application of the (fixed) function to its arguments.
    \item The interpretation of datatype relations use the (fixed) datatype relation itself.
    \item The interpretation of the equality predicate is the identity relation.
    \item Counting quantifiers have the obvious interpretation.
  \end{itemize}
\end{definition}

\subsection{eMAPL rules}

Because we are interested in effective reasoning, we restrict axioms to rules.
We make the usual additions and allowances for datatype predicates and functions.

\begin{definition}
  An eMAPL rule is a MAPL rule modified as in eMAPL.
  (Note that the functions in MAPL rules are different from datatype functions.)
  A variable is relational if it occurs in the rule body as the argument of a predicate
  that is not a datatype relation (including the datatype equality relations).
  Terms in rule bodies can use datatype functions, but the variables in these terms
  must be relational.
  Atoms in rules can use datatype relations but variables in these atoms 
  must be relational.
\end{definition}

Remember that in MARPL ontologies attributes are processed using rules
that take attributes and their values from instantiations of body atoms and
determine how they augment the head instantiation that is the consequent of
the rule.  We want to be able to handle attributes (Wikidata qualifiers) in
a uniform manner so that each rule does not need to say, for example, how
the Wikidata temporal qualifiers are processed.

We do this by characterizing each Wikidata qualifier, providing instructions
on how it is to be processed in each rule that does not say anything about
how the attribute's values are added to the consequences of the rule.

\begin{definition}
  An attribute is characterized in one of several ways.
  \begin{itemize}
  \item No values for the attribute are to be added to the consequents of
    rules.  We expect that most qualifiers will fit into this
    {\em ignore} characterization, so it is the default.
  \item The values of the attribute in the facts matched by body atoms are
    each added to the consequents of rules, in an {\em additive} fashion.
  \item The consequents of rules are given a single value for the
    attribute, formed by {\em combining} all the values for the attribute in
    matched body atoms by a datatype function that maps pairs of values in
    one datatype into a value in the same datatype.  If the result fails to
    satisfy a unary predicate then the rule produces no result.  
  \item The most sophisticated characterization involves starting by
    combining values for several attributes as above and then {\em blending} the
    resultant values using a different function, provided that the resultant
    values satisfy a datatype relation.  
  \end{itemize}
\end{definition}

Combination is only suitable for an attribute whose value has the datatype of
the function.  The datatype function should be commutative and associative,
but other functions could be used if there is only a single value for the
attribute, with the resultant value being the left reduction of the values
taken in the order of occurence in the body of the rule.

\noindent
\textbf{Wikidata qualifiers.}
    Wikidata has the qualifier \wdt{point in time}, which is used to state
    when some statements are valid.  Combining statements with this
    qualifier should take into account whether the time intervals overlap,
    so this qualifier would be characterized as a combining qualifier with
    the intersection function and the emptiness predicate.

    Other Wikidata statements use the qualifiers \wdt{start time} and \wdt{end time} to
    indicate an extended period when the statement is valid.  To
    produce the value for \wdt{start time} when making inferences from statements with
    these qualifiers requires first finding the last possible start time and
    the first possible end time and then taking the part of the combined
    start times that can be before the combined end times.

    So the Wikidata \wdt{start} qualifier could be 
    % handled by first
    characterized as a blending qualifier by first
    combining \wdt{start} and \wdt{end} qualifiers using the \wdt{last} and
    \wdt{first} functions, respectively, giving start and end times for the
    result, and then using the \wdt{could\_be\_before} function to cut out
    start times that cannot be before the end time.  If there is no start
    time before the end time (\wdt{not\_could\_be\_before} is true between
    the combined start and end times), then the inference produces no
    result.\footnote{In actuality, all three of these qualifiers should take into account all
    of them, so a better characterization would have a more complex blending
    specification.}

\begin{definition}
  An eMARPL ontology consists of a finite set of eMAPL rules, a finite set of function definitions, and a finite set of attribute characterizations as above.
  The function definitions are as in MARPL ontologies except that they can have datatype function terms 
  in their consequences (i.e., in the consequents of their conditionals).
\end{definition}

These attribute characterizations are used as macros, modifying the
functions and rules of an eMARPL ontology.

\begin{definition}
  The expansion of an eMARPL ontology uses the attribute characterization to modify its rules as follows:
  \begin{itemize}
    \item First give each rule its own function by making copies of
      functions or creating a new function for the rule.  Ensure that each
      atom in the rule body has its own set variable, adjusting the function
      as needed to access this set variable.
    \item For each attribute whose characterization is addition, for each rule
      where the function does not mention the attribute in its consequences,
      augment the function to copy over all the attribute's values.  
    \item For each attribute whose characterization is combine,
      for each rule where the function does not mention the attribute in its
      consequences, augment the function to add a single attribute value
      that is the combination of all the values in the body atoms.  Also add
      a new body atom to check whether the combined value satisfies the
      combination predicate.  This will require adding multiple clauses to
      the function as well as splitting the rule to take into acccount the
      presence or absence of the attribute in body atoms.  
    \item For each attribute whose characterization is blend, for each
      rule where the function does not mention the attribute in its
      consequences, augment the function to compute the combination values
      and add the blend result.  Also add a new body atom to check whether
      the blended value satisfies the blend predicate.  This will cause
      similar but more pronounced increase in the number of clauses than
      combination attributes.
  \end{itemize}
\end{definition}

The meaning of an expanded eMARPL ontology on an eMARS is just the inferential closure of the rules on the eMARS. 

\subsection{Complexity}

An expansion can be exponentially larger than the original ontology in the
number of attributes with combination or blending characterizations.
Implementations would not actually construct the expansion, instead
gathering up values internally and applying the functions and predicates
only to the existing values.  So the size of the expansion is not a real
complexity problem, by itself.

Datatype functions and relations are limited so that rules cannot add
new datatype values to predicate extensions, eliminating this particular
cause of intractability or undecidability in rules.

Even so, reasoning in MARPL, let alone eMARPL, is intractable  because the number of attribute values can grow large.  We view this as unavoidable, but we do not expect this worst-case intractability to be much of a problem in practice.

\section{Wikidata on eMARS}
\label{sec:Wikidata-on-eMARS}

Now that eMARPL ontologies have been defined we can show how we determine
the meaning of Wikidata as an eMARS.  Following that, we will show how the eMARS can be used to check Wikidata's constraints.

Objects in Wikidata are {\em items}, which include predicates ({\em properties}).
Facts in Wikidata are {\em statements}, consisting of a {\em subject} (an item) and a main {\em snak}.  
Snaks are predicate-object pairs, or {\em some-value} snaks, or {\em no-value} snaks.
Statements also have associated {\em qualifiers}, which are also snaks.
Statements have a rank, which is {\em regular}, {\em preferred}, or {\em deprecated}.
Wikidata also provides optional typing information for the values of properties.
We also use a characterization for each property used in a qualifier in
Wikidata; a set of ontological rules for Wikidata; and a set of constraints.

First we turn Wikidata itself into an eMARS.  
The domain elements of the eMARS are all the items in Wikidata, with additions as below.
The predicate extensions in the eMARS are pairs of subjects and objects from statements with the predicate as property.
Statements whose main snak is a some-value snak are modelled by adding a fresh element to the domain.
(This treats a some-value snak as stating that its value is a distinct domain element, which might not be quite the best treatment, but avoids computational problems having to do with determining which regular domain element the value is equal to.)
Each qualifier snak becomes an attribute of the statement, in the obvious manner.
For statement ranks we add a special rank attribute with value preferred or normal.
We ignore reference records for now.\tr{\footnote{One way to handle reference records is to add new objects to the domain for reference records and make these objects be values of a special qualifier, such as \wdt{reference\_record}.}}

Statements whose main snak is a no-value snak are ignored here, and left for future work\tr{, because the meaning of no-value snaks is uncertain in Wikidata}.
We think the best treatment of a no-value snak is as a constraint but it is unclear whether a no-value snak means no value at all, no value with the same qualifiers, or something in between.  These options can be modelled as eMAPL constraint formulae.

Property typing is modelled as rules requiring that the values of the property belong to the datatype.  So the datatype for \wdt{date\_of\_birth} is modelled by the rule:
\begin{equation}
  \begin{aligned}
    \wdt{date\_of\_birth}(s,o) \rightarrow \wdt{Point\_in\_time}(o)
  \end{aligned}
\end{equation}
\vspace*{-5ex}

\subsection{Wikidata Ontology Rules}

Some parts of formalizing the Wikidata ontology have already been
described above as examples of quantifier characterizations.
Arguably, the most important part of this formalization, though, is
formalizing the ontological rules of Wikidata.  

The ontology problems with Wikidata can be solved by treating Wikidata as an eMARPL
ontology, transforming Wikidata into an eMARS and adding a few ontology rules, such as those shown in (\ref{eqn:onto-axioms}).
Each atom on the right-hand side of these rules is implicitly associated with an attribute set that is constructed by a rule function.  
\vspace*{-0.5ex}
\begin{equation}
\label{eqn:onto-axioms}
  \begin{aligned}
  \textsf{\small subclass\_of}(c,d) \land \textsf{\small subclass\_of}(d,e) &\rightarrow \textsf{\small  subclass\_of}(c,e)   \\
  \textsf{\small instance\_of}(y,c) \land \textsf{\small subclass\_of}(c,d) &\rightarrow \textsf{\small instance\_of}(y,d) \\
  \textsf{\small subproperty\_of}(c,d) \land \textsf{\small subproperty\_of}(d,e) &\rightarrow \textsf{\small subproperty\_of}(c,e) \\
  \textsf{\small subproperty\_of}(p,q) \land p(x,y) &\rightarrow q(x,y) \\
  \textsf{\small instance\_of}(p, \textsf{\small symmetric\_property}) \land p(y,x) &\rightarrow p(x,y) \\
  \textsf{\small instance\_of}(p, \textsf{\small transitive\_property}) \land p(x,y)\land p(y,z) &\rightarrow p(x,z)
  \end{aligned}
\end{equation}

These rules look higher order, but they are not.  The quantification is only over Wikidata properties, 
\trs{so the rules can be rewritten using a triple-based formulation, as in:}{
so the fourth rule, for example, is more properly written as
\begin{displaymath}
  \textsf{\small instance\_of}(p,\textsf{\small Wikidata\_property}) \land \textsf{\small subproperty\_of}(p,q) \land p(x,y) \rightarrow q(x,y)
\end{displaymath}
Even the appearance of being higher order can be removed by changing to a triple-based formulation, as is done in RDF, where the rule would be written something like:}
\vspace*{-0.5ex}
\begin{displaymath}
  \textsf{\small T}(p,\textsf{\small instance\_of},\textsf{\small Wikidata\_property}) \land
  \textsf{\small T}(p,\textsf{\small subproperty\_of},q) \land
  \textsf{\small T}(x,p,y)
  \rightarrow 
  \textsf{\small T}(x,q,y)
\end{displaymath}

So far, these ontology rules are just MARPL, and even just regular Horn rules.  But
Wikidata qualifiers need to be taken into account, even if qualifiers are
forbidden in ontology atoms (i.e., those with predicates \textsf{\small instance\_of},
\textsf{\small subclass\_of}, \textsf{\small subproperty\_of}), because of occurrences of non-ontology atoms such as $p( x, y)$ in some rules.  In
MARPL, such rules would either have to take into account every possible
Wikidata qualifier, or different rules would have to be written for each
Wikidata property, but even these rules would have to take into acount all
the qualifiers that are present on Wikidata statements for the property.
In eMARPL a single rule can be written, without regard to qualifiers, and
the qualifier characterizations are used to handle the qualifiers that are
present.

\subsection{Wikidata Qualifier Characterization}

The most important, and most involved, Wikidata qualifier characterizations
are probably the ones for temporal qualifiers, as they impact very many
statements in Wikidata and missing temporal qualifiers can lead to
considering statements that are not relevant to the user's context.  The
characterizations of temporal qualifiers above as blending attributes are, we feel, typical and
show off the capabilities of eMARPL.

Many Wikidata qualifiers have little logical import or are specific to a
statement, such as the measurement method used to determine a statement or
the location of an event.  The former can be simply ignored for most uses of
Wikidata and the latter, although of interest, should not be carried along
in most inferences, so they would be classified in the default
classification.  Other Wikidata qualifiers have logical import and thus
should be carried along, but are multivalued and do not in general interact
with other qualifiers.  These can be categorized in the ``additive''
categorization.

\subsection{The Meaning of Wikidata}

The meaning of Wikidata is then the inferential closure of the eMARS above
under this eMARPL ontology.  It is this eMARS that is to be used when
querying or otherwise requesting what is true in Wikidata, or checking constraints.

\subsection{Constraints}

Wikidata has constraints that do not add directly to the meaning of
Wikidata but instead provide signals that there
is something questionable in Wikidata, 
consistent with the view taken by other work on constraints for knowledge-graph-like systems \cite{shacl,shex}.
We model Wikidata constraints as eMAPL
formulae that are evaluated over the eMARS that is the meaning of an eMARPL
ontology. 
\tr{We note that attribute sets and functions for checking set membership, which are useful for expressing constraints involving qualifiers, are already included in the definition of MAPL in \cite{marx2017logic}, and we have also included equality and counting quantifiers in eMAPL.
pfps removed for space considerations:
Constraints can either be given a positive formulation, which indicates a pattern of data elements that conform to the constraint, or a negative formulation, which indicates a pattern of data elements that violate the constraint.  In our view, it is most natural to first write the positive formulation, and then from that derive the negative formulation, which can then be used as a query.}
Instantations of these constraints that are false are to be reported as violations of the constraint.

For example, the \textsf{\small distinct\_values\_constraint} in Wikidata is supposed to say that a given property should have different values for different items.
Constraints like this one are currently implemented (if at all) by special-purpose code.  The following eMAPL formula embodies this constraint
\begin{equation}
\label{eqn:distinctValuesConstraint-a}
\begin{aligned}
\textsf{\small property\_constraint}(p, \textsf{\small distinct\_values\_constraint}) \land \\
p(s1, o1) \land p(s2, o2) \land s1 \neq s2 
\rightarrow o1 \neq o2
\end{aligned}
\end{equation}

\tr{Formula \ref{eqn:distinctValuesConstraint-a} directly expresses the meaning of the constraint in the usual fashion of first-order logic.  When this formula is satisfied for all bindings of the free variables the constraint holds.  Bindings of the free variables that falsify the formula are violations of the constraint.
The query formulation of this constraint is just its negation:
\begin{equation}
\label{eqn:distinctValuesConstraint-b}
\begin{aligned}
\textsf{\small property\_constraint}(p, \textsf{\small distinct\_values\_constraint}) \land \\
p(s1, o1) \land p(s2, o2) \land s1 \neq s2 
\land o1 = o2
\end{aligned}
\end{equation}
Bindings of the free variables that make Formula \ref{eqn:distinctValuesConstraint-b}
true, i.e., the results of the query, are violations of the constraint.}

In a separate paper \cite{martin2020logical}
we show that with the proposed formalism nearly all of Wikidata's existing property constraints can be given a complete characterization in an economical, natural, and relatively easy-to-understand fashion.  Such characterizations, unlike documentation in natural language, provide an unambiguous basis for understanding and for implementing constraint checkers.  
\tr{Constraint violations can be found by an expression-evaluator  for the formalism, with no additional engineering effort.  In other words, all that is required is to obtain the constraint's negative formulation (which, as mentioned earlier, can be automatically derived from its positive formulation) and submit it to the evaluator.}
Moreover, we show that additional constraints can easily be added, so long as they are expressible in eMAPL.  eMAPL allows for representing and handling a broad range of constraints, which goes beyond property constraints, in the same formalism.   

\subsection{Rules for Recognizing Class Instances}

There are many classes in the Wikidata ontology that have very few instances, even when considering instances of subclasses.  For example, as of mid-June 2020 in Wikidata \textsf{\small female\_human} has only 5 instances, even including instances of subclasses, as opposed to the several million expected.  The problem is that most female humans are stated as belonging to human and having \textsf{\small sex\_or\_gender} \textsf{\small female}.  There is nothing in Wikidata to suggest, however, that asking for instances of \textsf{\small female\_human} is unreasonable.

Adding a rule of the form
\begin{equation}
  \textsf{\small instance\_of}(\textsf{\small human}, h) \land \textsf{\small sex\_or\_gender}(h,\textsf{\small female})
  \rightarrow \textsf{\small female\_human}(h)
\end{equation}
results in \textsf{\small female\_human} having the correct instances.

Many other notable under-populated classes can be handled in the same way.

\section{Related Work}

We highlight relevant work from several slices of logical foundations for knowledge bases, 

\emph{Logical foundations for Wikidata.}
SQID \cite{marx2017sqid} is a browser and editor for Wikidata, which draws inferences from a collection of MARPL rules\footnote{SQID's rule set may be viewed at https://tools.wmflabs.org/sqid/\#/rules/browse.}.  Our work was informed by SQID's embodiment of MARPL-based reasoning, and motivated in part by the desire to expand the expressiveness of MARPL rules, as illustrated by the SQID rule set 
\tr{(particularly limitations related to attribute sets)} 
to provide a more complete reasoning framework, and to accommodate Wikidata constraints.
 \cite{hanika2019discovering} also formalizes a model of Wikidata based on  MARS, but with a different objective: the application of ``Formal Concept Analysis to efficiently identify comprehensible implications that are implicitly present in the data''.  \cite{hanika2019discovering} is thus nicely complementary with \cite{marx2017logic} and with our work, in that it provides a basis for discovering, rather than hand-authoring, new (e)MARPL rules. 
 
\emph{Logical foundations for annotated KBs.} 
Annotated RDFS \cite{zimmermann2012general} extends RDFS and RDFS semantics
to support annotations of triples. A
deductive system is provided, and extensions to the SPARQL query language that enable querying of annotated graphs.  While this approach could provide a useful target formalism for Wikidata's RDF dumps, we have chosen instead to represent Wikidata's data model as directly as possible, and thus we deliberately avoid the use of the RDF dumps, and the complexities that could arise from adopting RDF as the modeling framework.

\emph{Adding attributes to logics.}
Just as MARPL was developed to provide a (rule-based, Datalog-like) decidable fragment of MAPL, Krötzsch, Ozaki, et al. have also explored description logics as a basis for other
decidable fragments of MAPL, and have analyzed the resulting family of attributed DLs in \cite{krotzsch2017attributed,krotzsch2017reasoning,krotzsch2018attributed,ozaki2018happy,ozaki2019temporally}.  We believe that MARPL provides the best available starting point for modeling Wikidata, but we also agree that this ongoing thread of research will lead to attributed DLs with the right level of expressivity for other sorts of applications.

\section{Conclusion and Future Work}

We have described eMARS, eMAPL, and eMARPL (extensions of previous work of \emph{Marx et al.} \cite{marx2017logic}), which can provide a practically usable foundation for a logic of Wikidata.  These extensions are centered around the introduction of datatypes and the specification of the behavior of qualifiers in connection with axioms.  We have outlined a reasoning framework for using these formalisms with Wikidata, and have indicated how its use could add substantial value to Wikidata\tr{ and in particular bring benefits from well-founded reasoning involving axioms, constraints, qualifiers, and special values}.  
The adoption of this framework could support more complete, meaningful, and consistent querying of Wikidata, as well as better facilities for implementing KB completion and other reasoning capabilities.

Our plans for future work include the
discussion of other areas of Wikidata that can benefit from its use
and the creation of a prototype implementation and then a detailed design for a scalable deployment.

\bibliographystyle{splncs} 
\bibliography{main}

\end{document}